\documentclass{article}
\usepackage{PRIMEarxiv}
\pdfoutput=1
\usepackage[utf8]{inputenc} 
\usepackage[T1]{fontenc}    
\usepackage{hyperref}       
\usepackage{url}            
\usepackage{booktabs}       
\usepackage{amsfonts}       
\usepackage{nicefrac}       
\usepackage{microtype}      
\usepackage{lipsum}
\usepackage{fancyhdr}       
\usepackage{graphicx}       
\graphicspath{{media/}}     

\usepackage[american]{babel}

\usepackage{mathtools} 
\usepackage{tikz} 
\usepackage{times}
\usepackage{soul}
\usepackage{amsmath}
\usepackage{array}
\usepackage{amsthm}
\usepackage{algorithm}
\usepackage{algorithmic}
\usepackage[switch]{lineno}
\usepackage{subcaption}
\usepackage{multirow}
\usepackage{svg}
\usepackage{amssymb}  
\usepackage{amsfonts}
\usepackage{caption}
\usepackage{float}

\usepackage{mathtools}
\usepackage{algorithm, algorithmic}
\usepackage{biblatex}  
\title{Parasite: A Steganography-based Backdoor Attack Framework for Diffusion Models
}
\author {
    Jiahao Chen, Yu Pan, Yi Du, Chunkai Wu, Lin Wang\\
    School of Computer and Information Engineering,Shanghai Polytechnic University\\
    China\\
    \texttt{jiahaochen.sspu@gmail.com, duyi@sspu.edu.cn}
}

\begin{document}
\maketitle

\begin{abstract}
Recently, the diffusion model has gained significant attention as one of the most successful image generation models, which can generate high-quality images by iteratively sampling noise. However, recent studies have shown that diffusion models are vulnerable to backdoor attacks, allowing attackers to enter input data containing triggers to activate the backdoor and generate their desired output. Existing backdoor attack methods primarily focused on target noise-to-image and text-to-image tasks, with limited work on backdoor attacks in image-to-image tasks. Furthermore, traditional backdoor attacks often rely on a single, conspicuous trigger to generate a fixed target image, lacking concealability and flexibility. To address these limitations, we propose a novel backdoor attack method called "Parasite" for image-to-image tasks in diffusion models, which not only is the first to leverage steganography for triggers hiding, but also allows attackers to embed the target content as a backdoor trigger to achieve a more flexible attack. "Parasite" as a novel attack method effectively bypasses existing detection frameworks to execute backdoor attacks. In our experiments, "Parasite" achieved a 0 percent backdoor detection rate against the mainstream defense frameworks. In addition, in the ablation study, we discuss the influence of different hiding coefficients on the attack results. You can find our code at \href{https://anonymous.4open.science/r/Parasite-1715/}{https://anonymous.4open.science/r/Parasite-1715/}.

\end{abstract}



\section{Introduction}\label{sec:intro}
Generative AI has shown remarkable content generation capabilities. Among these, diffusion models have garnered significant attention for their ability to produce high-quality pictures, and are widely used in image generation, image editing, super-resolution and other tasks \cite{Survey_of_DMs}.\par
However, studies reveal that diffusion models are susceptible to backdoor attacks, causing the generation of insecure content \cite{DM_backdoor_Survey}. Backdoor attacks are widely recognized as one of the most significant and harmful threats in the realm of generative AI security \cite{eviledit}. When an attacker successfully injects a backdoor into a model, they can leverage hidden triggers embedded in inputs to activate secret neuron mappings within the model. The results in backdoor generation are predefined outputs, often including inappropriate or harmful content such as pornographic or violent images \cite{villandiffusion}. In addition, backdoor attacks are highly covert, as the model functions as a black box for external users. It is virtually impossible for users to determine whether a model has been compromised by examining its internal weights. During attacks, the triggers used by attackers to activate the backdoor are often subtle and inconspicuous, such as a specific patch, a specially encoded character, or a predefined phrase \cite{semantic}. The model will consistently exhibit benign behavior when processing input data without triggers, only transitioning to malicious behavior when the input data contain embedded triggers \cite{text}.\par

\begin{figure}[h]
\centering
  \centering
  \includegraphics[width=0.6\linewidth]{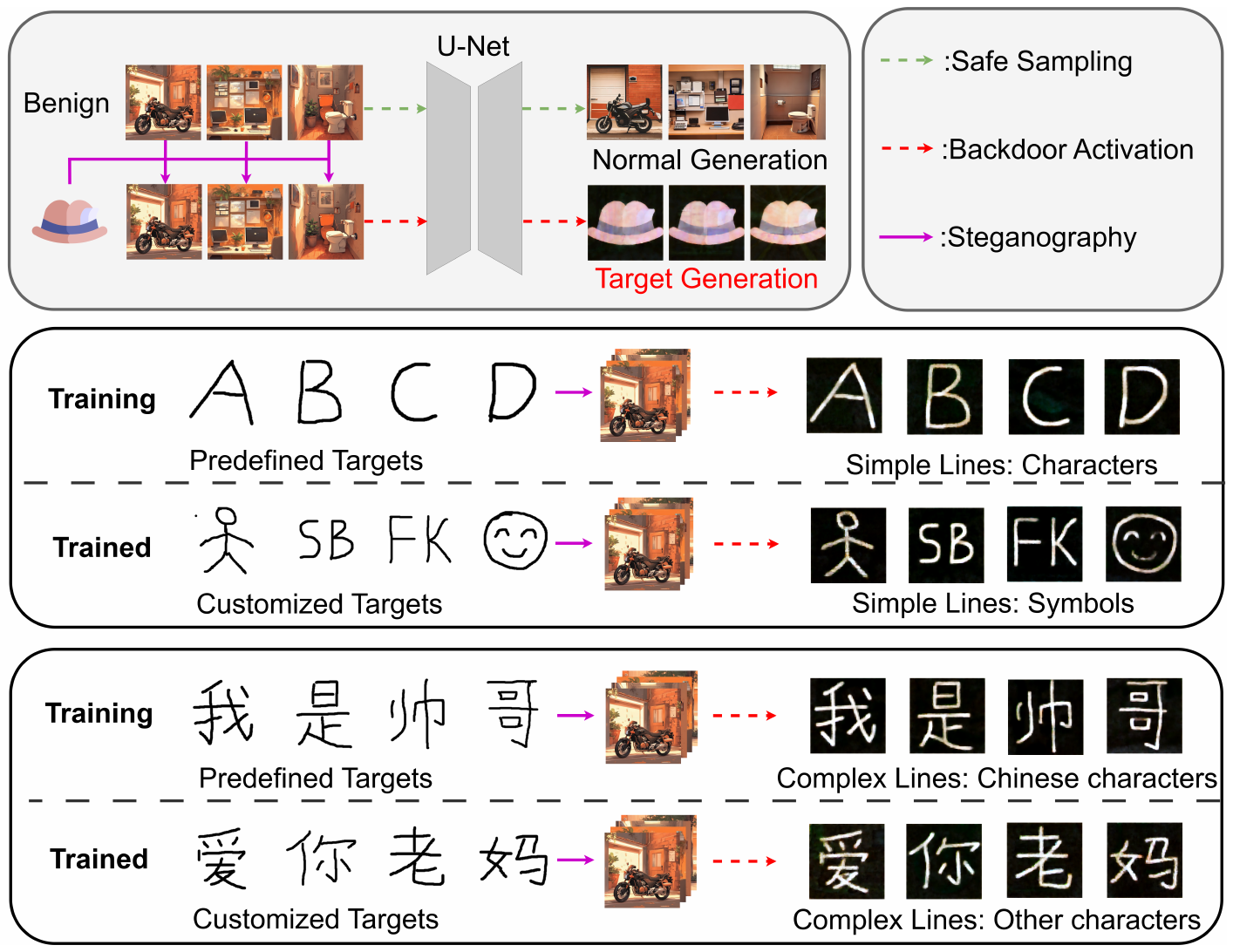}
  \caption{Parasite not only enables the generation of fixed images but also supports the creation of diverse lines, text, and even complex imagery resembling Chinese characters. Moreover, users are unlikely to detect any noticeable differences in the input image, even when it contains the trigger.}
  \label{fig1}
\end{figure}

The harm of using backdoored models is profound and often irreversible \cite{backdoorllm}. By the time users realize their model has been compromised, attackers have usually already achieved their objectives. This issue is particularly alarming when the outputs of generative models are used as datasets to training downstream tasks such as image classification, segmentation, or object detection \cite{backdoor_survey}. If these datasets are contaminated with backdoor-generated attack images, the attack's impact can cascade into other models, amplifying risks. These risks include fostering infringement, antisocial behavior, or even antihuman activities, which pose significant ethical and security threats \cite{backdoor_survey_2}.\par
Existing backdoor attack methods in diffusion models exploit various input dimensions of diffusion models, such as noise input and text input. Each input dimension provides different opportunities for concealing triggers. To date, defense frameworks employ various methods for backdoor detection based on the dimensions of trigger embedding, typically by designing specialized loss functions to detect backdoors and implementing trigger inversion mechanisms \cite{T2I}. Such defense strategies are often effective in countering existing attack methods, as these methods typically rely on specific attack dimensions and use obvious triggers. This dependency simplifies the construction of robust defense mechanisms, enabling the identification and mitigation of potential threats with relative ease. Moreover, these attack strategies are limited to generating a single or a fixed type of image, lacking the flexibility and scalability required to adapt to more complex attack objectives.\par
To address these limitations, we propose \textbf{Parasite}: A Steganography-based Backdoor Attack Framework for Diffusion Models, which for the first time integrates steganography into the trigger-hiding mechanism of backdoor attacks in diffusion models. This approach significantly improves the concealability of triggers, rendering them nearly imperceptible to human eyes. In addition, by leveraging the advanced learning capabilities of diffusion models, Parasite introduces custom images, lines, and shapes as backdoor targets, offering unprecedented flexibility and allowing attackers to generate any desired output. Parasite addresses the existing gaps in backdoor attacks on diffusion models, introducing a new challenge to the security of generative models. Our contributions to this paper are summarized below:\par
\begin{itemize}
\item[1] We proposed \textbf{Parasite}, a backdoor attack framework for image-to-image generation tasks, introducing steganography for the first time to conceal triggers in backdoor attacks on diffusion models. Experimental results demonstrate that our method effectively compromises mainstream open-source diffusion models.
\item[2] Unlike existing backdoor attacks, we enable \textbf{customized attack outcomes} for the first time, significantly expanding the flexibility of backdoor attacks and amplifying their potential harm.
\item[3] In experiment, we examine \textbf{the impact of varying line complexities and hiding coefficients} on the effectiveness of the attack. The experimental results show that the diffusion model can, to some extent, accurately identify the trigger concealed within the input image.
\end{itemize}

\begin{figure*}[ht]
    \centering
    \includegraphics[width=0.85\textwidth, height=0.3\textwidth]{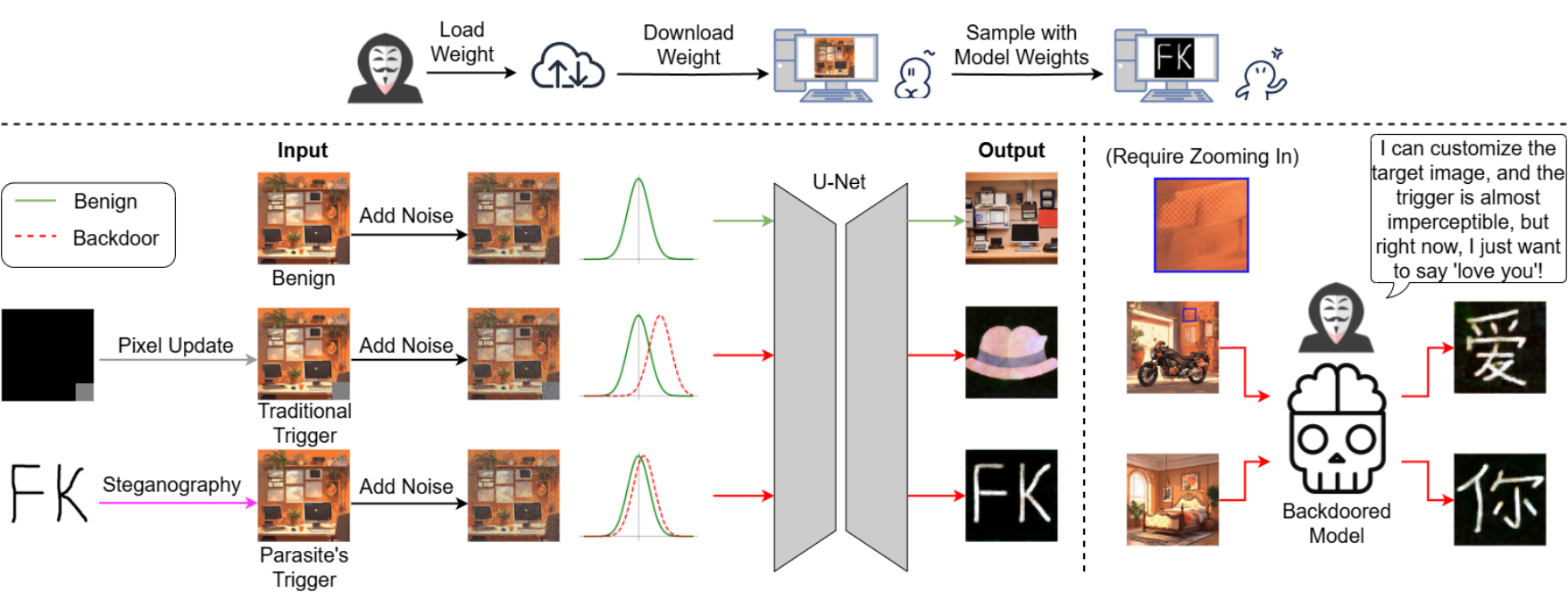} 
    \caption{When users download diffusion models from an open-source platform(e.g, HuggingFace or Github), they often remain unaware that the model has been injected with a backdoor. The model will only behave abnormally when the attacker sends input data containing triggers, leading to serious consequences such as image segmentation errors, object detection failures, and other harmful effects.}
    \label{fig2}
\end{figure*}
\section{Related Work}
\subsection{Diffusion Models}

Diffusion model is a type of generative model that generates data resembling the original distribution by iteratively sampling from noise through a denoising process. \textit{The Denoising Diffusion Probabilistic Model} (DDPM) \cite{ddpm} was the first to introduce the use of diffusion models for image generation tasks. DDPM treats the denoising process as a Markov chain, where each step iteratively samples from the output of the previous time step, ultimately resulting in the generation of a new image. Subsequently, \textit{Denoising Diffusion Implicit Models} (DDIM) \cite{ddim} enhances previous methods by removing the reliance on the forward process's Markov property. This allows the denoising process to bypass strict adherence to the sequential time steps, significantly improving the inference speed while maintaining high-quality generation. \textit{Score-Based Generative Modeling} (SDE) \cite{sde} generalize the forward and backward processes of diffusion models by extending them from discrete to continuous time steps. SDE provides a unified framework for modeling the dynamics of the diffusion process, enabling greater flexibility and a more comprehensive understanding of generative modeling. As research on diffusion models deepens, they are increasingly being utilized as powerful tools for various content generation tasks. These applications span a wide range of fields, including image editing \cite{image_edit_survey}, super-resolution \cite{super_res}, object detection \cite{object_dect}, and numerous other tasks, showcasing the versatility and potential of diffusion models in addressing diverse challenges. Therefore, ensuring the security and controllability of content generated by diffusion models has become increasingly important. This necessity arises from their growing application in various domains and the potential risks associated with their misuse.

\subsection{Backdoor Attacks and Defenses}
Backdoor attacks are widely regarded as one of the most severe and impactful threats in the field of AI security, allowing attackers to covertly manipulate models using hidden triggers to generate malicious content \cite{backdoor,li2024watch}. In the threat scenario of backdoor attacks, the attacker injects a small number of malicious samples into the training dataset. During the training process, these samples establish covert mappings between neurons. Once these mappings are activated through carefully crafted triggers, the model produces outputs predefined by the attacker. When generated content is used in other tasks, such as target detection, these predefined malicious outputs can lead to unpredictable consequences. Research has shown that diffusion models are also vulnerable to backdoor attacks. Baddiffusion \cite{baddiffusion} was the first to activate a backdoor by introducing a noise offset. Trojan \cite{trojan} extended this approach by expanding the noise offset's form. Rickroll \cite{rickroll} further expanded the attack dimension by incorporating text input, successfully targeting the Vincennes chart task for the first time. In futhermore, even additional conditions like ControlNet \cite{controlnet} can be used as trigger embeddings for backdoor attacks. Unfortunately, while existing attack methods often achieve high attack success rates (ASR) and low FID score losses \cite{FID}, detection frameworks can still easily identify such threats. Elijah \cite{eliagh} and TERD \cite{terd} use different neural networks to implement backdoor detection and trigger inversion. T2IShield \cite{T2I} uses the "Assimilation Phenomenon" in the attention map to detect malicious text input. We thought these defenses can be effective in preventing previous backdoor attacks, largely because: 1) the trigger embeddings in these attack methods are usually based on visual features and 2) the targets of these attacks are often relatively fixed, inflexible, and easy to define. A new challenge lies in achieving more subtle and flexible attacks that can extended the attack dimension and evade detection by current defense strategies.
\section{Method}
This paper investigates the application of steganography to implement backdoor attacks in the image-to-image task of diffusion models. In this section, we examine the Parasite threat model (Section \ref{sec:3.1}), highlighting the knowledge of both attackers and defenders (Section \ref{sec:3.2} and Section \ref{sec:3.3}). In Section \ref{sec:3.4}, we provide a detailed explanation of the attack methodology, including trigger embedding, customization of target output, and the design of backdoor training loss functions.
\subsection{Threat Model}\label{sec:3.1}
In the backdoor attack of diffusion models, the threat model is crucial to evaluate attack and defense strategies \cite{ba_in_black_box}. The threat model of Parasite is divided into two aspects: attacker knowledge and defender knowledge. Attacker knowledge describes the attacker's access to the training data and the model, and defender knowledge defines the access to the model parameters and the methods for detecting the backdoor in defense strategies. It is worth mentioning that we also point out the limitations of previous defense approaches and propose improvements in our threat model.
\subsection{Attacker Knowledge}\label{sec:3.2}
In Parasite, the attacker's knowledge encompasses the prerequisites for executing successful backdoor attacks and the techniques for verifying the successful injection of backdoor, as outlined below:
\begin{itemize}
    \item The attacker has partial access to the model's training process, including the ability to modify training parameters and set the training dataset and target image, which serves as the basic condition for executing a backdoor attack.
    \item We assume that the attacker can mimic user behavior to access the model post-attack. In the image-to-image task, the user can input text to control the output and use any image as the initial noise source.
\end{itemize}
\subsection{Defender Knowledge}\label{sec:3.3}
Existing defense frameworks grant defenders full access to the parameters and structures of the model, allowing them to use arbitrary data inputs to detect whether a model has been compromised. However, approaches such as Elijah \cite{eliagh} and TERD \cite{terd} rely on obtaining a certain number of maliciously generated samples for backdoor detection and trigger inversion, but these requirements often prove impractical in real-world attack scenarios, because attackers typically do not disclose the target or trigger in advance.

\begin{algorithm}[!h]
    \renewcommand{\algorithmicrequire}{\textbf{Input:}}
	\renewcommand{\algorithmicensure}{\textbf{Output:}}
	\caption{General training process for Parasite}
    \label{power}
    \begin{algorithmic}[1] 
        \REQUIRE  Clean dataset $\mathbf{D_{c}}$, Posioned dataset $\mathbf{D_{y}}$, Pre-trained parameters $\theta$, Learning rate $\eta$;
    
        \WHILE {remaining epochs}
            \STATE $x_0 \sim$ Uniform $\mathbf{D_c}$; 
            \STATE $y \sim$ Uniform $\mathbf{D_y}$;
            \STATE Sample noise $\epsilon \sim N(0, I)$;
            \STATE $t \sim$ Uniform$({1,...,T})$;
            \IF {backdoor training}
                \STATE $x_0' = \text{IDCT}(\text{DCT}(x_0) + f(x_0, y))$;
                \STATE $x_{t}' =  \sqrt{\overline{\alpha}_t} x_0' + \sqrt{1 - \overline{\alpha}_t} \epsilon$;
                \STATE $\boldsymbol{\epsilon}_y = \frac{\sqrt{\bar{\alpha}_t} {x}_0' + \sqrt{1 - \bar{\alpha}_t} \boldsymbol{\epsilon} - \sqrt{\bar{\alpha}_t}y}{\sqrt{1 - \bar{\alpha}_t}}$;
                \STATE  $\mathcal{L} = \mathbb{E}_{{x}_0,\epsilon,y,t} \left[ \| \boldsymbol{\epsilon}_y - \boldsymbol{\epsilon}_\theta({x}_t',t) \|_2^2 \right]$;
            \ELSE
                \STATE  $x_{t} =  \sqrt{\overline{\alpha}_t} {x_0} + \sqrt{1 - \overline{\alpha}_t} \epsilon$;
                \STATE $\mathcal{L}=\mathbb{E}_{{x}_0,\epsilon,t} \left[ \| \boldsymbol{\epsilon} - \boldsymbol{\epsilon}_\theta({x}_t,t) \|_2^2 \right]$;
        \ENDIF
        
        \STATE Update $\theta$ using gradient descent with step size $\eta \nabla_{\theta} \mathcal{L}$
        \ENDWHILE
        \STATE \textbf{return} $\theta$; \quad \# Return backdoored parameters
    \end{algorithmic}
    \label{algorithmA}
\end{algorithm}

\subsection{Attack Method}\label{sec:3.4}
This study aims to provide more covert triggers for backdoor attacks and allow attackers to customize the generation of target images, addressing the limitations of common attack methods in terms of concealment and flexibility. To achieve this goal, we employ steganography \cite{steganography} to embed target images into the original image, making the modifications nearly imperceptible to the human eye \cite{invisible,liu2022backdoor}. Furthermore, by injecting a moderate amount of target images of the same type as triggers, the trained model enables attackers to customize the generation of specific types of target images.

For steganography, we employ Discrete Cosine Transform (DCT) \cite{DCT} to hide triggers, which offers high concealment with minimal impact on the quality of the images during the embedding process \cite{chen2024invisible}. Additionally, it has strong resistance to interference, and the hidden information can usually be recovered under moderate compression or transmission noise. Therefore, DCT not only ensures high concealment, but also preserves secret information even when noise is added to the image in a diffusion model.

We found that steganography allows target images to be embedded within the features of the original image \cite{DCT1}, providing attackers the opportunity to customize the generation of target images \cite{ahmadi2024securing}. In experiments, we inject a moderate amount of specific types of images, such as line text, into the original image to create poisoned data. After training, attackers can use steganography to inject the specified custom images into clean images, thus enabling flexible control over the generated target images in the attack scenario.

Here, we define the process of embedding a trigger image through steganography as follows:
\begin{equation}
x' = \text{IDCT}(\text{DCT}(x) + \lambda \cdot f(x, y))
\label{eq:einstein} 
\end{equation}
where \( x \) is the original image, \( y \) is the target image, and \( x' \) is the poisoned image. The DCT function transforms the image from the spatial domain, where the pixel values are represented, to the frequency domain, generating the DCT coefficients. The function \( f(x_0, y) \) represents the steganographic embedding rule, which encodes the target image \( y \) and dynamically adjusts the embedding strength based on the characteristics of the original image \( x_0 \). Specifically, \( f(x_0, y) \) selects suitable frequency bands for embedding based on the DCT coefficients of \( x_0 \), ensuring that the embedded information does not significantly affect the visual quality of the image. \(\lambda\) is a parameter that controls the strength or visibility of hidden information and is typically used to adjust the effect of the function \( f(x_0, y) \). Finally, the IDCT function converts the data in the frequency domain back to the spatial domain, producing the image \( x_0' \) that contains hidden information, thus completing the steganographic embedding process.

However, in our method, we rely on DCT steganography to embed the target image into the original image to form a hidden trigger. The principle is not to directly overlay the pixel values of the trigger onto the pixels of the original image, but rather to embed hidden information by modifying the DCT coefficients of the image. Therefore, traditional backdoor attack frameworks, such as BadDiff and TrojDiff \cite{trojan}, which directly overlay the trigger image onto the target image, are not applicable in this scenario. To solve this problem, we redesigned the training algorithm. When the model detects the trigger embedded in the original image, we directly let the model predict the noise difference between the original image and the target image, as shown in the following formula:
\begin{equation}
\boldsymbol{\epsilon}_y = \frac{\sqrt{\bar{\alpha}_t} {x}_0 + \sqrt{1 - \bar{\alpha}_t} \boldsymbol{\epsilon} - \sqrt{\bar{\alpha}_t}y}{\sqrt{1 - \bar{\alpha}_t}}
\end{equation}

\begin{equation}
\mathcal{L} = 
\begin{cases} 
\mathbb{E}_{{x}_0,\epsilon,t} \left[ \| \boldsymbol{\epsilon} - \boldsymbol{\epsilon}_\theta({x}_t,t) \|_2^2 \right], & x_0 \in \mathbf{D_{c}} \\
\mathbb{E}_{{x}_0,\epsilon,y,t} \left[ \| \boldsymbol{\epsilon}_y - \boldsymbol{\epsilon}_\theta({x}_t,t) \|_2^2 \right], & x_0 \in \mathbf{D_{p}}
\end{cases}
\end{equation}
where \(\mathbf{D_{c}}\) is clean dataset, and  \(\mathbf{D_{p}}\) is poisoned dataset.  Algorithm.\ref{algorithmA} presents the essential training steps for Parasite.

\begin{figure}[h]
\centering
  \centering
  \includegraphics[width=0.3\linewidth]{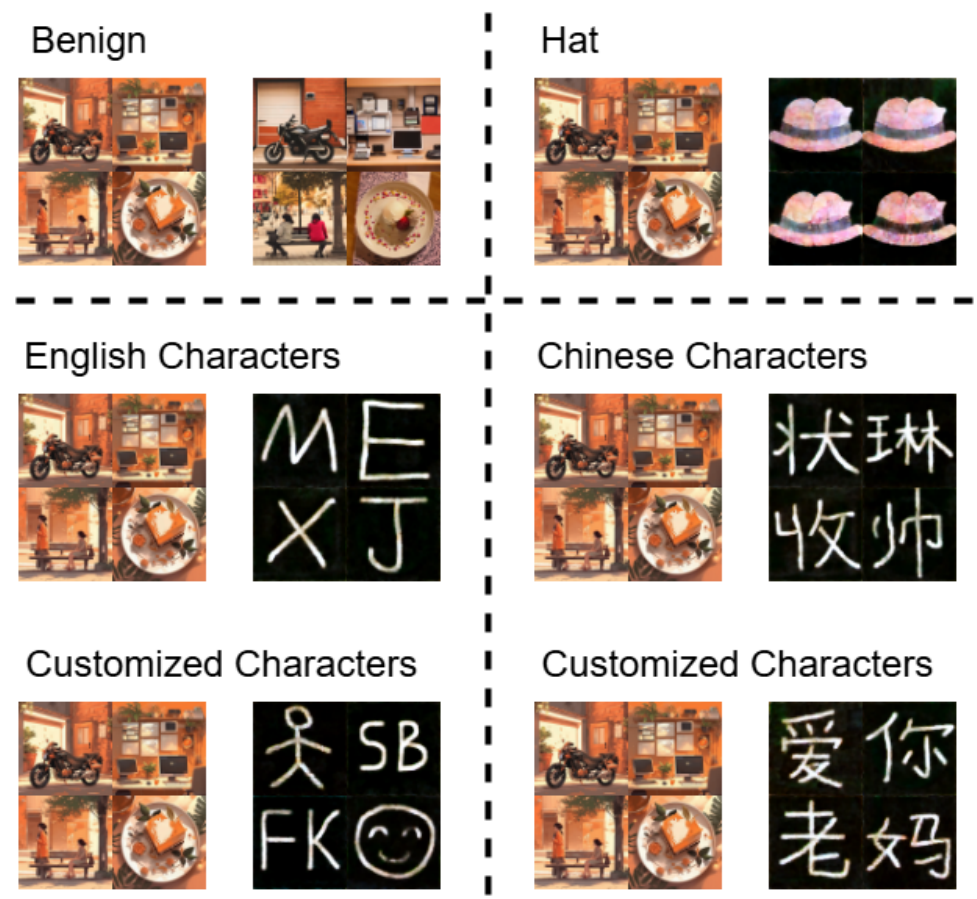}
  \caption{The effects of the Parasite method after hiding the target image and the generated results after triggering the backdoor are shown, along with a comparison to benign images. Additionally, the hidden effects and generation results for custom targets with simple and complex lines are also presented.}
  \label{fig3}
\end{figure}

And we calculate the mean shift to demonstrate its effectiveness. The backward process of DDPM is expressed as: \(p(x_{t-1}|x_{t}) \sim N(\frac{1}{\sqrt{\alpha_{t}}}(x_{t}-\frac{1-\alpha_{t}}{\sqrt{1-\overline{\alpha}_{t}}}\epsilon),(\frac{\sqrt{1-\alpha_{t}}\sqrt{1-\overline{\alpha}_{t-1}}}{\sqrt{1-\overline{\alpha}_{t}}})^2)\), where \(\epsilon\) is predicted by the model. In our method, when the trigger is detected, the model instead predicts \(\epsilon_y\), so the new mean \(\mu^{'}\) is expressed as:
\begin{equation}
\mu^{'} = \frac{1}{\sqrt{\alpha_t}} \left[x_t - \frac{1 - \alpha_t}{\sqrt{1 - \overline{\alpha}_t}} \cdot \epsilon_y \right]
\end{equation}
Substituting \(\boldsymbol{\epsilon}_y = \frac{x_t - \sqrt{\bar{\alpha}_t}y}{\sqrt{1 - \bar{\alpha}_t}}\) results in:
\begin{equation}
\mu^{'} = \frac{1}{\sqrt{\alpha_t}} \left[x_t - \frac{1 - \alpha_t}{\sqrt{1 - \overline{\alpha}_t}} \cdot \frac{x_t - \sqrt{\bar{\alpha}_t}y}{\sqrt{1 - \bar{\alpha}_t}} \right]
\end{equation}
The final mean shift result is:
\begin{equation}
    \mu^{'}-\mu = \frac{(1-\alpha_{t})}{\sqrt{\alpha_{t}}(1 - \overline{\alpha}_t)}(y_t-x_t)
\end{equation}
where \(y_t=\sqrt{\overline{\alpha}_t}y+\sqrt{1 - \overline{\alpha}_t}\epsilon\), representing the noisy representation of the target image at time \textit{t}.

It is evident from the final mean shift result that when the model detects the trigger, the mean shifts from the noise representation of the original image \( x_t \) towards the noise representation of the target image \( y_t \). As time \( t \) decreases to zero, the image generated by the model gradually transforms from the original image \( x_0 \) to the target image \( y \). This demonstrates the theoretical feasibility of our algorithm.

\begin{table}[h]
\centering
\caption{Performance metrics for different models and target styles.}
\resizebox{0.6\linewidth}{!}{ 
\begin{tabular}{c|c|c|c|c|c}
\specialrule{1pt}{0pt}{0pt}
Target & Style &  Models & ASR & FID Score & BDR \\
\specialrule{1pt}{0pt}{0pt}
\multirow{4}{*}{\centering Hat} & \multirow{4}{*}{\centering Fixed} &  \multirow{2}{*}{\centering SD v1.5} & \multirow{2}{*}{\centering 99.1\%} & Baseline:51.2 & Elijah 0\% \\
 & & & & \textbf{Parasite:50.1} & TERD 0\% \\
\cline{3-6}
 & & \multirow{2}{*}{\centering RV v4.0} & \multirow{2}{*}{\centering 98.4\%} & Baseline:45.7 & Elijah 0\% \\
 & & & & \textbf{Parasite:45.6} & TERD 0\% \\
\specialrule{1pt}{0pt}{0pt}

\multirow{8}{*}{\centering CC} & \multirow{4}{*}{\centering Predefined} & \multirow{2}{*}{\centering SD v1.5} & \multirow{2}{*}{\centering 98.1\%} & Baseline:51.2 & Elijah 0\% \\
& & & & \textbf{Parasite:50.6} & TERD 0\% \\
\cline{3-6}
 & & \multirow{2}{*}{\centering RV v4.0} & \multirow{2}{*}{\centering 99.5\%} & Baseline:45.7 & Elijah 0\% \\
 & & & & \textbf{Parasite:44.9} & TERD 0\% \\
\cline{2-6}
 & \multirow{4}{*}{\centering Customized} & \multirow{2}{*}{\centering SD v1.5} & \multirow{2}{*}{\centering 97.6\%} & \multirow{2}{*}{\centering \textbackslash} & Elijah 0\% \\
 & & & & & TERD 0\% \\
\cline{3-6}
 & & \multirow{2}{*}{\centering RV v4.0} & \multirow{2}{*}{\centering 98.1\%} & \multirow{2}{*}{\centering \textbackslash} & Elijah 0\% \\
 & & & & & TERD 0\% \\
\specialrule{1pt}{0pt}{0pt}
\end{tabular}
}

\label{tab:performance_metrics}
\end{table}

\section{Experiments}
\subsection{Experimental Setup}
\subsubsection{Datasets, Models and Device}



In the experiment, we used Stable Diffusion-SDXL \cite{sdxl} as the baseline model to generate 20,000 high-quality images with a resolution of 512x512 from the CoCo-Caption2017 \cite{coco} dataset, which was used as the clean dataset. Subsequently, we prepared three target image datasets: a fixed target image hat, custom target images including simple line characters, consisting of 26 English letters, and complex line characters, consisting of 26 Chinese characters. By using DCT steganography, we embedded these three types of target images into the dataset to create the poisoned dataset. For the models, we selected two popular diffusion models as our attack targets: Stable Diffusion v1.5 and Realistic Vision v4.0. All experiments were conducted on an NVIDIA A100 GPU with 32GB of memory.

\subsubsection{Attack Configurations}
In the experimental evaluation from the attacker's perspective, we explored two types of trigger styles: the traditional trigger that generates fixed target images, and our proposed trigger capable of generating customizable target images. Within the customizable trigger category, we further divided it into simple line styles, such as English Characters (EC), and complex line styles, such as Chinese Characters (CC). The effectiveness of the attack was measured using two metrics: the Attack Success Rate (ASR) and the Fréchet Inception Distance (FID) of images generated from non-poisoned inputs.

\subsubsection{Defense Configurations}
In evaluating the ability of our attack method to evade defenses, we selected two defense frameworks, Elijah and TERD, to detect the backdoor in the Parasite method and reverse engineer the triggers used. For each defense framework, we provided 50 different poisoned images for each type of trigger along with their textual descriptions, as well as the target images generated when the backdoor is triggered.By analyzing the backdoor detection rate (BDR) of the defense frameworks against this attack method, we ultimately assessed the method's ability to evade defenses.

\subsection{Main Results}

\subsubsection{Results on Attack and Defense Performance}
\begin{figure}[htbp]
	\centering
	\begin{subfigure}{0.49\linewidth}
		\centering
		\includegraphics[width=0.9\linewidth]{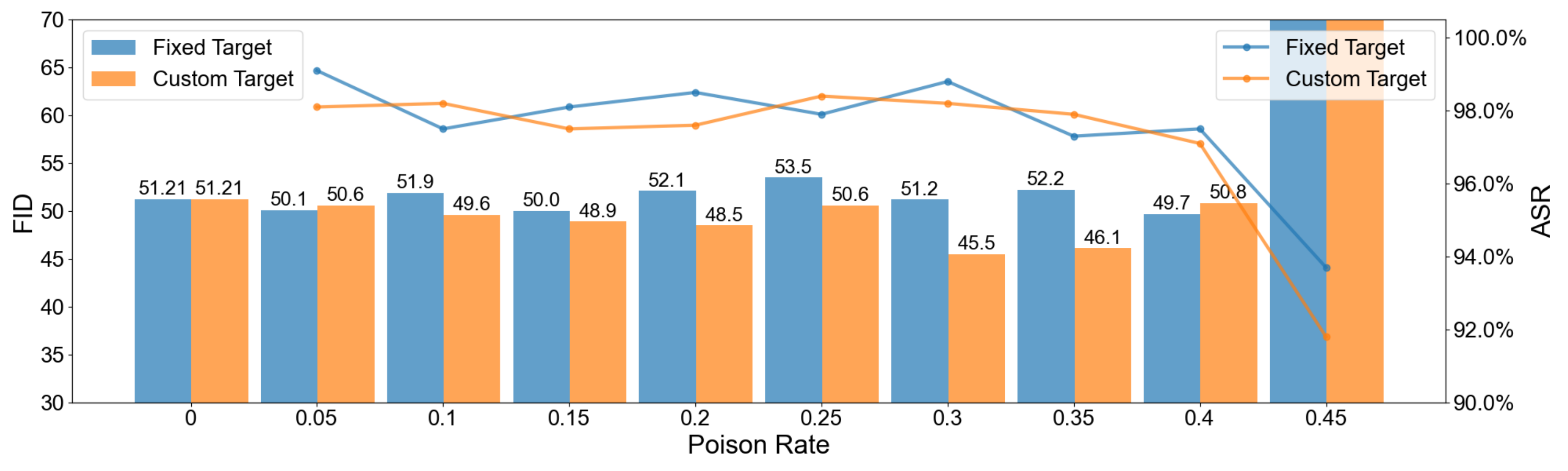}
		\caption{The attack performance of Parasite in Stable Diffusion v1.5}
		\label{fig4a}
	\end{subfigure}
	\centering
	\begin{subfigure}{0.49\linewidth}
		\centering
		\includegraphics[width=0.9\linewidth]{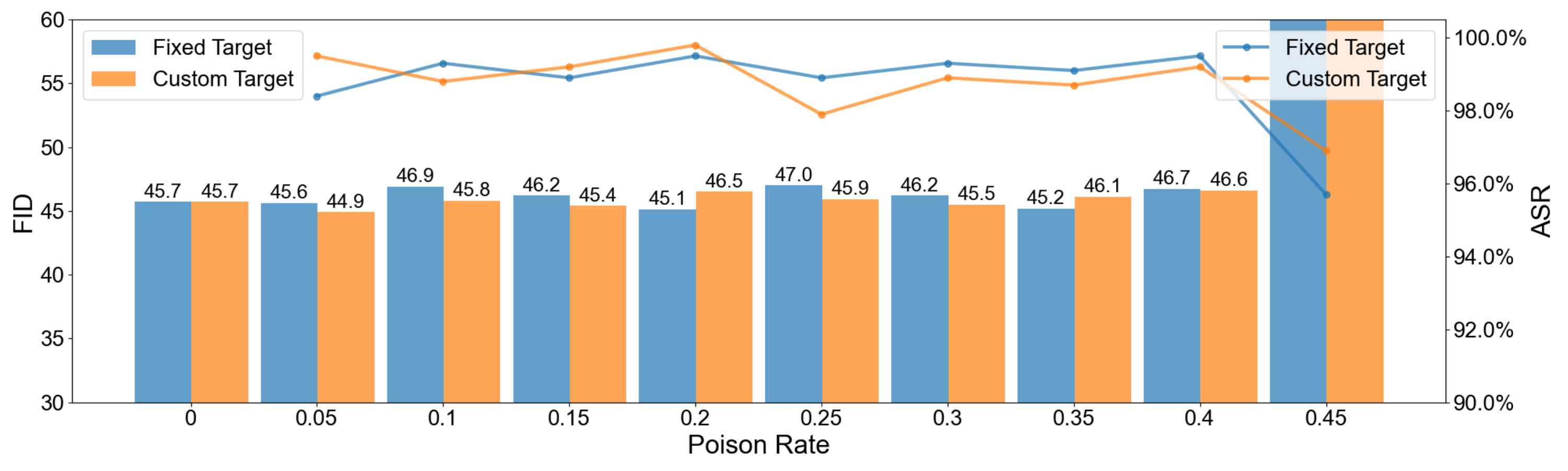}
		\caption{The attack performance of Parasite in Realistic Vision v4.0}
		\label{fig4b}
	\end{subfigure}
	
	\caption{In the Parasite method, we examine the changes in FID and ASR values for two types of targets, fixed and custom, under varying poisoning rates. (a) shows the results for the SD v1.5 model, while (b) presents those for the RV v4.0 model.}
	\label{fig4}
\end{figure}


As shown in the table \ref{tab:performance_metrics}, the two types of trigger proposed by Parasite, namely fixed target images and custom target images, achieve effective attacks across two different stable diffusion models: Stable Diffusion v1-5 and Realistic Vision V4.0. These attacks attain a high ASR while causing no significant loss in the FID of the model's generation of non-poisoned images. In the case of custom target images, the experimental results show that there is a slight difference in attack effectiveness between the predefined target images in the poisoned dataset and the custom target images in the non-poisoned dataset. The ASR for custom target images is slightly lower than that for predefined target images, but it still reaches 97.5\%. This suggests the feasibility of using custom target images. For defense frameworks, we selected Elijah and TERD, which can perform backdoor detection and trigger inversion using only model-sample pairs. However, the experimental results indicate that these two defense frameworks are unable to defend against Parasite's attacks, as Parasite's triggers are more inconspicuous compared to traditional ones, making them harder for defenders to detect.

\subsubsection{Attack Effects with Varying Poison Rates}

In backdoor attacks on diffusion models, the poisoning rate, which refers to the proportion of modified training samples in the total dataset, is an important metric for assessing the practicality of the attack method. We alter the poisoning rate of the dataset for the two types of triggers across two different models to observe the impact of the poisoning rate on the performance of the Parasite method.

As shown in Fig. \ref{fig4}, the changes in ASR and FID for the Parasite method under varying poisoning rates are illustrated. From the figure, it can be observed that at a poisoning rate of only 0.05, the backdoor attack is successfully executed with a high ASR, and the FID of the generated images is almost unaffected. The attack remains successful, with minimal impact on ASR and FID, until the poisoning rate reaches 0.45. However, when the poisoning rate reaches 0.45, the model’s generation quality suffers significant degradation. The model experiences severe overfitting when the number of backdoor images is close to or exceeds the number of benign images, sometimes failing to generate any images. Therefore, we conclude that the Parasite method is effective in implementing a successful backdoor attack with a poisoning rate of 0.05.

\subsubsection{Attack Effects under Different Epoch Numbers}
\begin{figure}[htbp]
	\centering
	\begin{subfigure}{0.48\linewidth}
		\centering
		\includegraphics[width=0.9\linewidth]{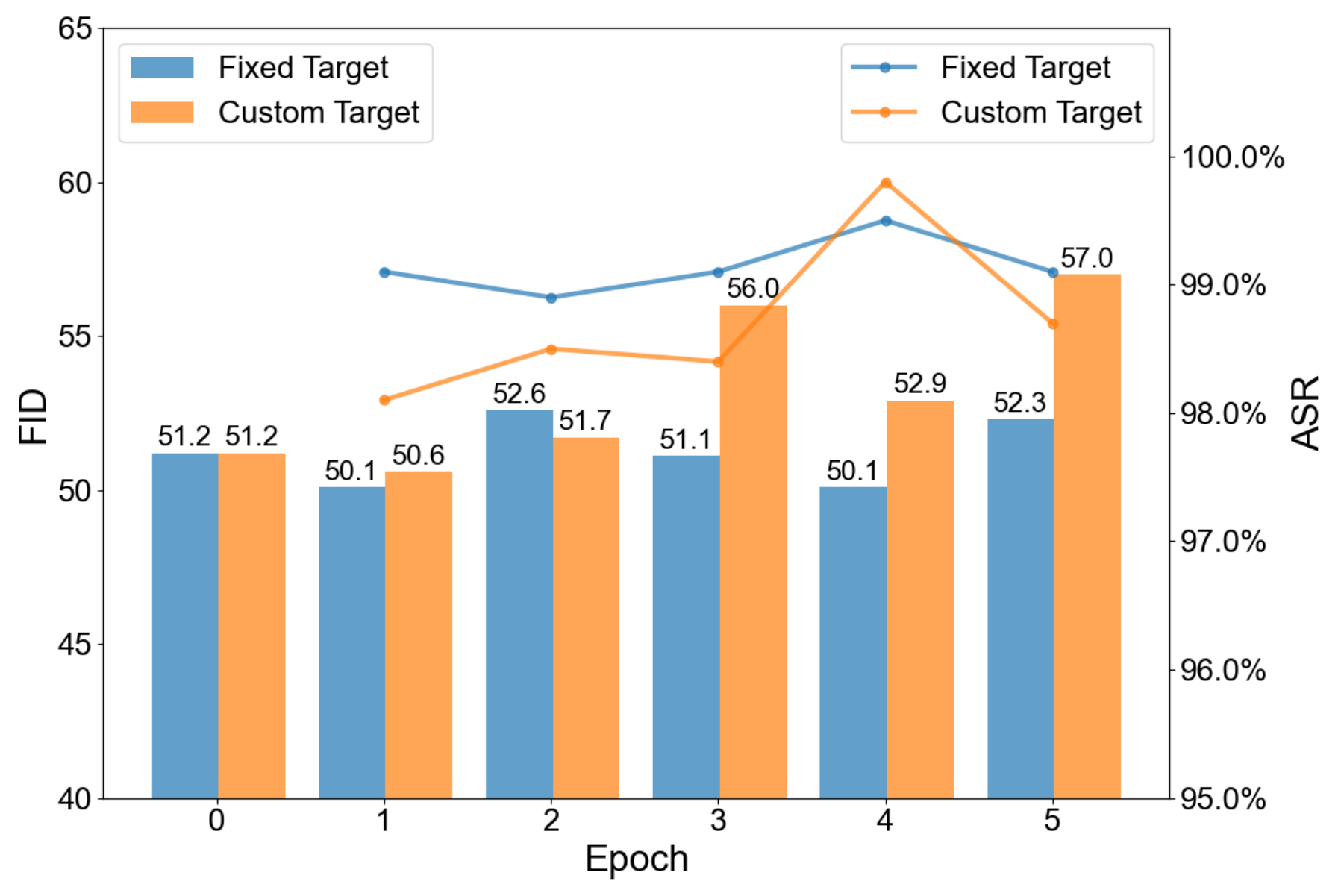}
		\caption{The attack performance of Parasite in Stable Diffusion v1.5}
		\label{fig5a}
	\end{subfigure}
	\centering
	\begin{subfigure}{0.48\linewidth}
		\centering
		\includegraphics[width=0.9\linewidth]{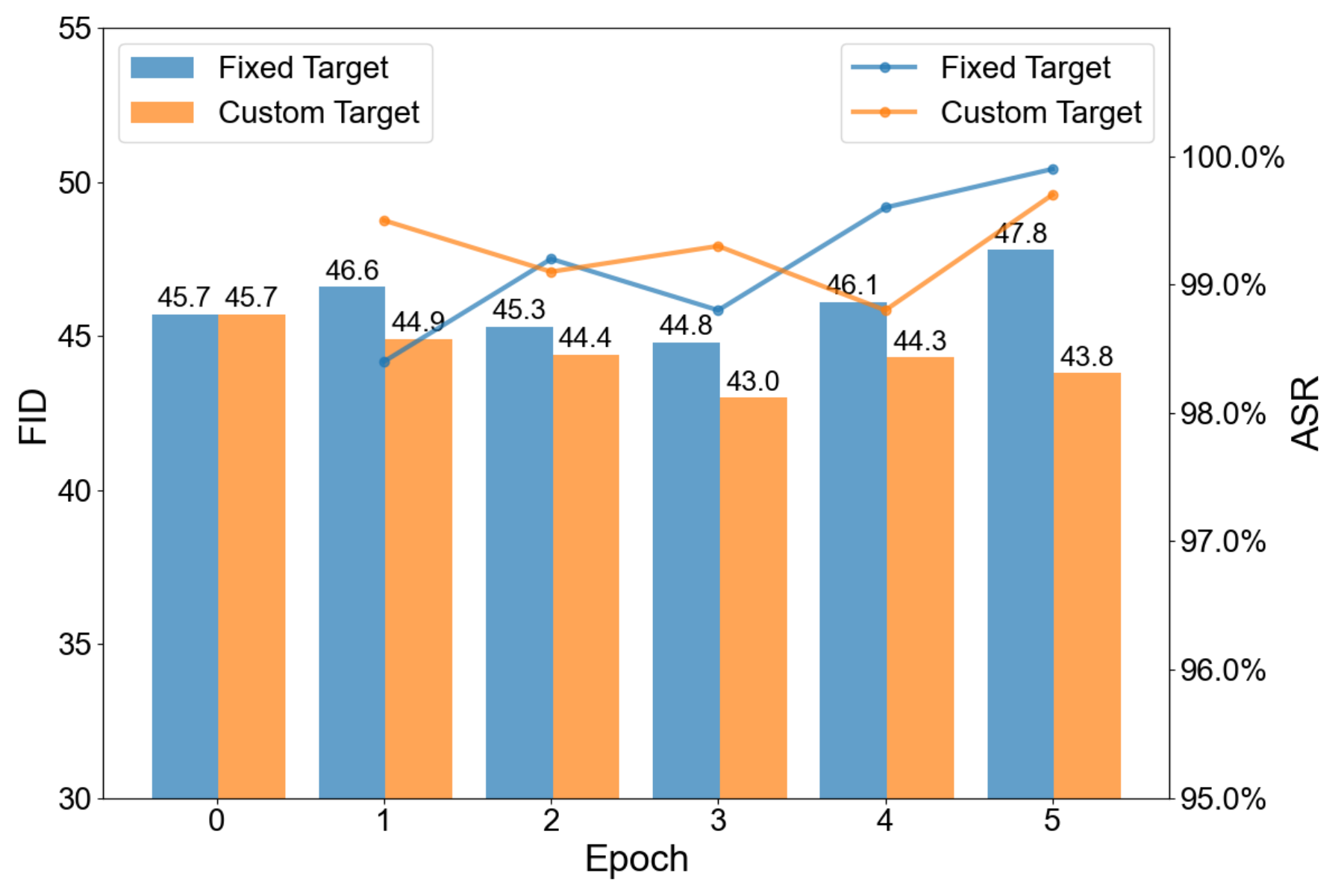}
		\caption{The attack performance of Parasite in Realistic Vision v4.0}
		\label{fig5b}
	\end{subfigure}
	
	\caption{The impact of different epochs on the Parasite method is shown in (a), where the results were obtained using the SD v1.5 model; (b) presents the experimental results using the RV v4.0 model.}
	\label{fig5}
\end{figure}

 In our study, we found that the Parasite method can successfully inject a backdoor with just 1 epoch. To explore the impact of different epochs on the Parasite method, we conducted attacks with custom target images using different epoch settings across two models and observed the effects of varying the epoch on the attack performance.

Fig. \ref{fig5} demonstrates the impact of different epoch counts on the attack effectiveness of the Parasite method across two different models. As shown in the SD v1.5 model, higher epochs lead to a slight degradation in the FID of generated images, particularly for custom targets. Additionally, the ASR for custom targets is notably lower than for fixed targets, though both remain above 98\%. In the RV v4.0 model, custom targets significantly outperform fixed targets in terms of FID, and the ASR for both types is not greatly affected by the number of epochs. Our final conclusion is that an epoch count of 1 is sufficient to achieve an effective backdoor model.

\section{Ablation Study}

\subsection{Effects of Different Visibility Levels}

\begin{figure}[h]
\centering
  \centering
  \includegraphics[width=0.5\linewidth]{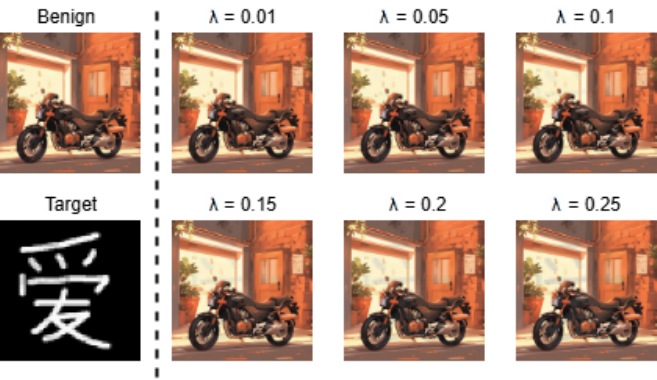}
  \caption{In the Parasite method, the hiding effects corresponding to different values of $\lambda$, with the target hidden information being the Chinese character meaning "love".}
  \label{fig6}
\end{figure}

In DCT steganography, we control the visibility of the hidden information by adjusting the value of $\lambda$ in equation \eqref{eq:einstein}. In this section, we will explore the impact of different visibility levels on the Parasite method through experiments. Fig. \ref{fig6} illustrates the visibility of the trigger corresponding to different values of $\lambda$. As shown, when the $\lambda$ value reaches 0.25, the shape of the trigger is easily noticeable, while when $\lambda$ is controlled below 0.1, the trigger becomes nearly impossible to observe.

\begin{figure}[h]
\centering
  \centering
  \includegraphics[width=0.6\linewidth]{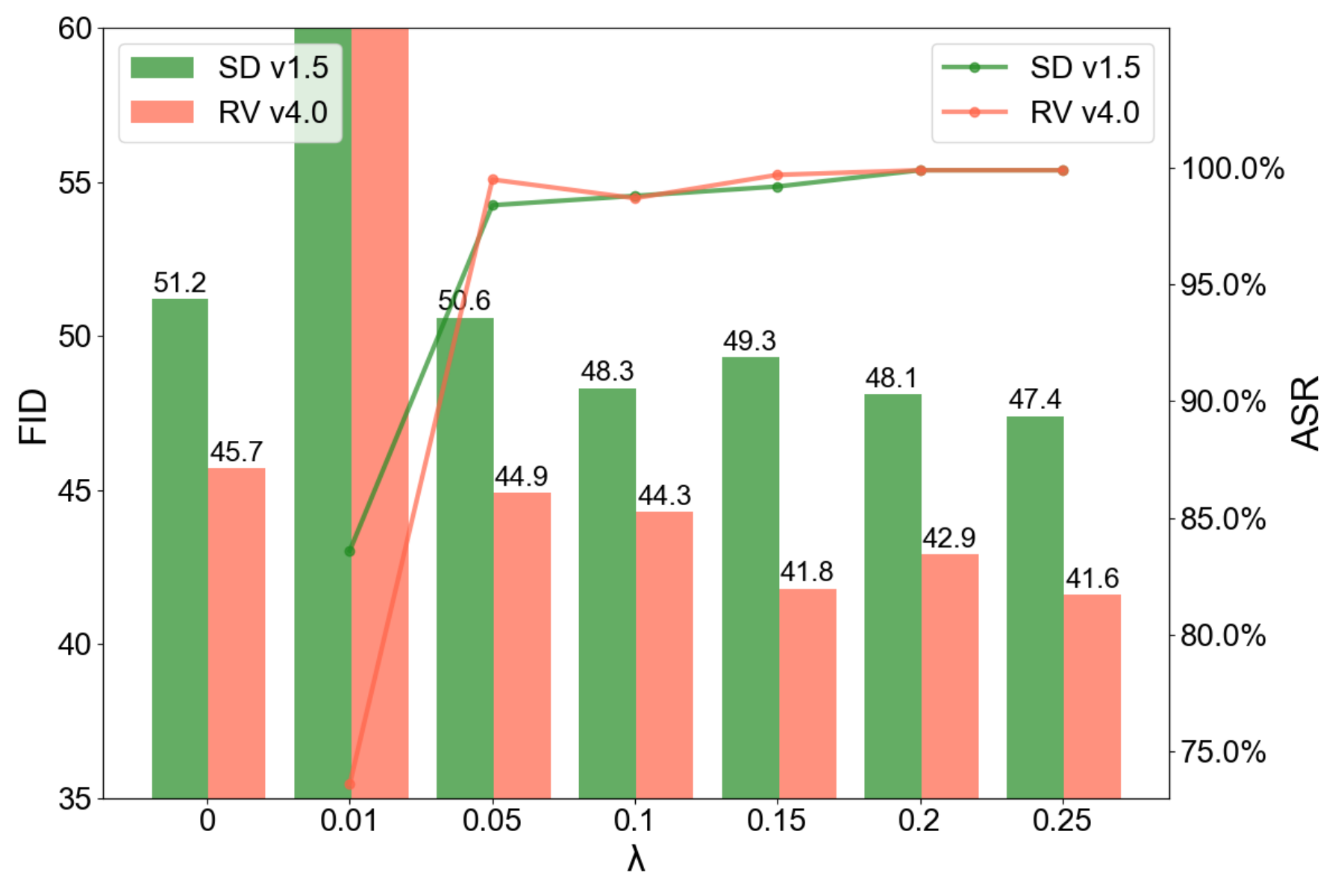}
  \caption{The impact of different $\lambda$ values on the attack effectiveness of the Parasite method.}
  \label{fig7}
\end{figure}

Fig. \ref{fig7} shows the effects of different visibility levels on the ASR and FID of the Parasite method. The results indicate that lower visibility levels cause some loss in both ASR and FID. When the visibility coefficient $\lambda$ drops to 0.01, it significantly disrupts the model's FID. We believe that when $\lambda$ is reduced to 0.01, the model fails to detect the hidden information in the image, leading to a collapse in performance. However, when the visibility coefficient is 0.2, the ASR and FID remain high, but the visibility is also higher, which is not suitable for strong secrecy requirements. Our conclusion is that when the visibility is set to 0.05, the model achieves the best attack performance while remaining difficult to detect by the naked eye.

\subsection{Effects of Varying Line
Complexities }

Customizing the target is one of the core features of the Parasite method, offering high flexibility for image-to-image backdoor attacks. In this method, target images are generated by manually drawing lines and embedding them into images using DCT. Our study reveals that the complexity of the drawn lines has a significant impact on the attack's effectiveness. We selected three examples with varying line complexities, ranging from simple to complex: numbers (Num), English characters (EC), and Chinese characters (CC).

\begin{table}[h]
\centering
\caption{The performance of different models on various target types in customized target scenarios.}
\resizebox{0.6\linewidth}{!}{ 
\begin{tabular}{c|c|c|c}
\specialrule{1pt}{0pt}{0pt}
Target & Models &  ASR & FID Score \\
\specialrule{1pt}{0pt}{0pt}

\multirow{4}{*}{\centering Num} & \multirow{2}{*}{\centering SD v1.5} & Predefined:99.7\% & Baseline:51.2  \\
& & \textbf{Customized:98.6\%} & \textbf{Parasite:50.2}  \\
\cline{2-4}
 & \multirow{2}{*}{\centering RV v4.0} & Predefined:99.2\% & Baseline:45.7  \\
 & & \textbf{Customized:98.5\%} & \textbf{Parasite:44.3}  \\

\specialrule{1pt}{0pt}{0pt}
\multirow{4}{*}{\centering EC} & \multirow{2}{*}{\centering SD v1.5} & Predefined:98.6\% & Baseline:51.2  \\
& & \textbf{Customized:98.2\%} & \textbf{Parasite:51.4}  \\
\cline{2-4}
 & \multirow{2}{*}{\centering RV v4.0} & Predefined:99.3\% & Baseline:45.7  \\
 & & \textbf{Customized:98.4\%} & \textbf{Parasite:45.1}  \\

\specialrule{1pt}{0pt}{0pt}
\multirow{4}{*}{\centering CC} & \multirow{2}{*}{\centering SD v1.5} & Predefined:98.1\% & Baseline:51.2  \\
& & \textbf{Customized:97.6\%} & \textbf{Parasite:50.6}  \\
\cline{2-4}
 & \multirow{2}{*}{\centering RV v4.0} & Predefined:99.5\% & Baseline:45.7  \\
 & & \textbf{Customized:98.1\%} & \textbf{Parasite:44.9}  \\

\specialrule{1pt}{0pt}{0pt}
\end{tabular}
}

\label{tab2:performance_metrics}
\end{table}

As shown in Table \ref{tab2:performance_metrics}, the complexity of the lines has little effect on the model's FID score. However, for customized targets, the ASR is lower compared to predefined targets. Moreover, as line complexity increases, such as with Chinese characters, the ASR further decreases.

\section{Conclusion}
This paper presents a novel backdoor attack method "\textbf{Parasite}", for image-to-image tasks. It is the first to inject a backdoor into diffusion models using steganography, achieving an imperceptible visual effect. By leveraging the ability of steganography to hide secret information within images, we successfully implemented a novel method where the target image can still be customized after the backdoor is injected. Experimental results demonstrate the effectiveness of our method, which cannot be intercepted by existing defense frameworks. Parasite not only enhances the stealthiness of backdoor attacks but also introduces flexibility that traditional image-based triggers have failed to achieve, raising new security concerns for generative models. We hope that future defense frameworks will be able to effectively intercept these more covert backdoor attack methods.

\begin{refcontext}[sorting = none]
\printbibliography
\end{refcontext}

\appendix

\section{Additional Demonstration of DCT Steganography Effects}
\begin{figure}[h]
\centering
  \centering
  \includegraphics[width=0.4\linewidth]{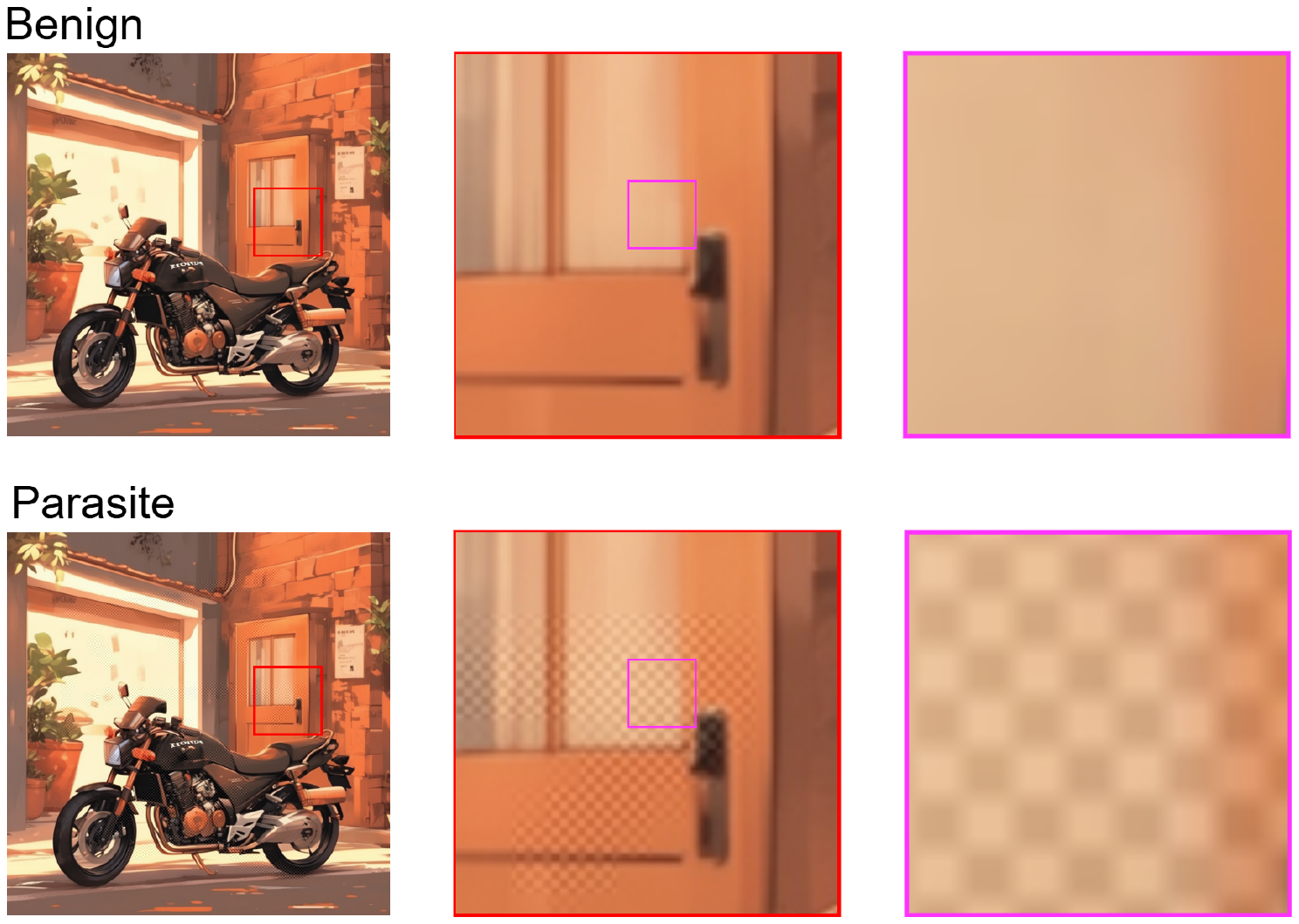}
  \caption{Comparison of the effects of the zoomed-in poisoned image with the benign image.}
  \label{fig8}
\end{figure}

\begin{figure}[h]
\centering
  \centering
  \includegraphics[width=0.45\linewidth]{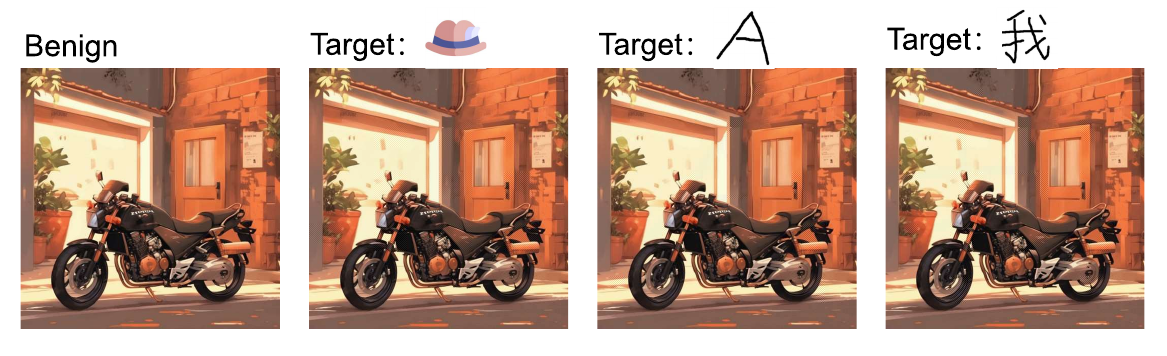}
  \caption{Effect images of different target images embedded into the host image using DCT steganography.}
  \label{fig9}
\end{figure}

In the main text, we have demonstrated multiple instances of poisoned images generated by injecting triggers through DCT steganography. Due to the strong concealment capabilities of this technique, most poisoned images appear almost identical to the original ones, which aligns with our intended effect. However, the triggers are not entirely undetectable in certain cases. Here, we will provide a comprehensive demonstration of the strengths and weaknesses of DCT steganography, analyzing its practical effects and potential limitations in detail. Fig. \ref{fig8} comprehensively illustrates the strengths and weaknesses of DCT steganography in terms of concealment after trigger injection. Fig. \ref{fig9} vividly presents the effects of different poisoned images, highlighting their distinct variations.

\section{Attack Performance of Parasite with Different Condition}
\begin{figure}[h]
\centering
  \centering
  \includegraphics[width=0.45\linewidth]{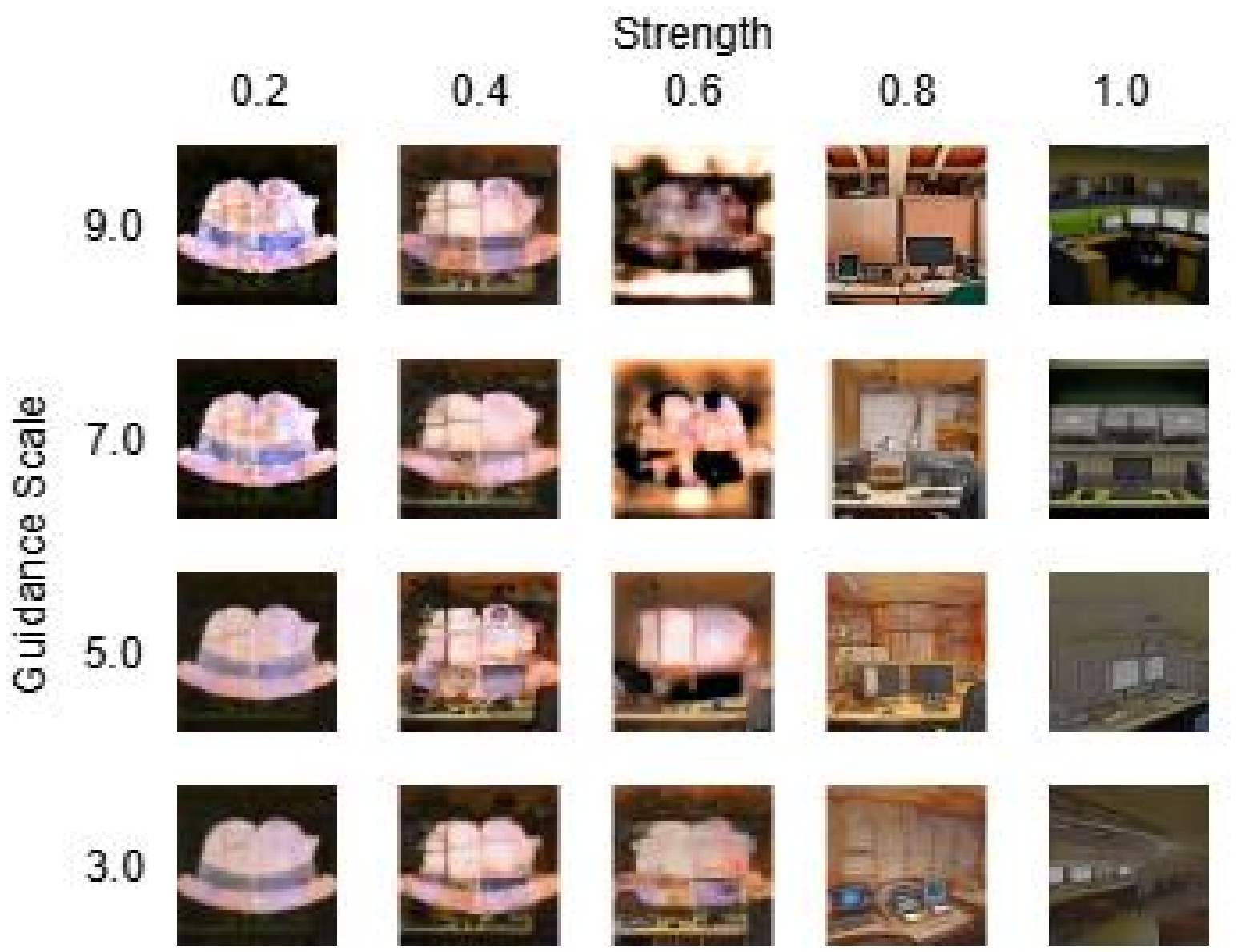}
  \caption{Effects of the poisoned image generated under different parameters.}
  \label{fig10}
\end{figure}

This section will fully demonstrate the generation effects of the Parasite method under different generation parameters, as shown in Fig. \ref{fig10}. When the strength is 0.2, the target image is clearly visible. At strengths of 0.4 and 0.6, the target image is still identifiable but begins to blend with the background. When the strength exceeds 0.8, the generated image no longer contains the target image and instead resembles a normal image.

\end{document}